\documentclass[10pt,a4paper,oneside]{article}

\usepackage{graphicx}
\usepackage{epstopdf}
\usepackage[noblocks]{authblk}
\usepackage{amssymb}
\usepackage{amsmath}
\usepackage{amsfonts}
\usepackage{mathtools}
\usepackage{amsfonts}
\usepackage{color}
\usepackage{listings}
\usepackage{fancyvrb}
\usepackage{float}
\usepackage{framed}
\usepackage{multirow}
\usepackage{hyperref}
\usepackage{url}  
\usepackage{multicol}
\usepackage{subcaption}
\usepackage{algorithm}
\usepackage{algpseudocode}
\usepackage{pifont}
\usepackage{algorithm}
\usepackage{colortbl}
\usepackage{framed}
\usepackage{listings}
\usepackage{bm}
\usepackage{eurosym}
\usepackage{amssymb}
\usepackage{suffix}
\usepackage{tikz}
\usetikzlibrary{matrix}

\usepackage[a4paper,left=2.75cm,right=2.75cm,top=3.5cm,bottom=3.5cm]{geometry}

\DeclarePairedDelimiterX\MeijerM[3]{\lparen}{\rparen}%
{\begin{smallmatrix}#1 \\ #2\end{smallmatrix}\delimsize\vert\,#3}

\newcommand\MeijerG[8][]{%
  G^{\,#2,#3}_{#4,#5}\MeijerM[#1]{#6}{#7}{#8}}

\WithSuffix\newcommand\MeijerG*[7]{%
  G^{\,#1,#2}_{#3,#4}\MeijerM*{#5}{#6}{#7}}

\begin{document}
\title{Optimal Counterfactual Explanations for Scorecard modelling}
\author{Guillermo Navas-Palencia}
\affil{\texttt{g.navas.palencia@gmail.com}}

\maketitle

\begin{abstract}
Counterfactual explanations is one of the post-hoc methods used to provide explainability to machine learning models that have been attracting attention in recent years. Most examples in the literature, address the problem of generating post-hoc explanations for black-box machine learning models after the rejection of a loan application. In contrast, in this work, we investigate mathematical programming formulations for scorecard models, a type of interpretable model predominant within the banking industry for lending. The proposed mixed-integer programming formulations combine objective functions to ensure close, realistic and sparse counterfactuals using multi-objective optimization techniques for a binary, probability or continuous outcome. Moreover, we extend these formulations to generate multiple optimal counterfactuals simultaneously while guaranteeing diversity. Experiments on two real-world datasets confirm that the presented approach can generate optimal diverse counterfactuals addressing desired properties with assumable CPU times for practice use.
\end{abstract}

\section{Introduction}

The use of interpretable machine learning models and the application of explainability methods to enhance the understanding of algorithmic decisions from black-box models have become crucial due to the widespread use of machine learning for decisions impacting society. Among the available explainability methods, counterfactual explanations \cite{Wachter2018} is one of the most promising for explaining predictions of black-box machine learning models by generating human-understandable post-hoc explanations. Counterfactual explanations provide information valuable to determine the changes required to the input variables to modify the outcome of a decision-making system without revealing the underlying algorithmic details. The changes, although possibly drastic, are more realizable when only a few variables require minor modifications. Thus, a practical counterfactual should be simple and as close as possible to the input data while obtaining a change on the desired outcome.

An important assumption for counterfactual explanations to be useful is that the underlying model does not differ when a user enters the system after a first unsatisfactory outcome. If this assumption holds, a counterfactual guarantees the desired outcome upon altering the input variables appropriately. 

From a fairness perspective, counterfactuals might be useful as a fairness evaluator to provide evidence that decisions generated by a system may be discriminatory.

Explainable machine learning has been of interest to the healthcare (e.g., medical diagnosis) and finance (e.g., lending) communities. Most examples in the literature, address the problem of generating post-hoc explanations after the rejection of a loan application. These examples tend to use black-box machine learning models, while in practice, decision-making systems for consumer finance in the banking industry employ globally interpretable scorecard models approved by banking regulators.

\subsection{Related work}
Existing optimization-based approaches for generating counterfactual explanations starts with the seminal work by Watcher et al. \cite{Wachter2018}. We focus on this type of methods and refer to \cite{Verma2020} for a complete and comprehensive review of alternative approaches. The work in \cite{Wachter2018} solves the Lagrangian relaxation of the general optimization problem using gradient-based methods, whereas other approaches \cite{Russel2019, Ustun_2019} solve the problem while providing a certificate of optimality or infeasibility when the underlying estimator is expressible in mathematical modelling terms, thus solvable by classical optimization algorithms. Furthermore, the latter two works develop procedures to generate diverse counterfactuals by iteratively adding new constraints. A more general approach generating diverse counterfactuals using gradient-based methods is studied in \cite{Mothilal_2020}. To generate realistic counterfactuals, the authors in \cite{Kanamori_2020} include new terms in the objective function to account for the empirical data distribution. Finally, in the context of multi-objective optimization for counterfactual explanations, we note the work in \cite{Dandl_2020}, where an evolutionary strategy is proposed to generate multiple counterfactuals while devising the Pareto frontier.

\subsection{Contributions}
In this work, we develop several mathematical optimization formulations to generating a single or multiple counterfactual explanations for a binary, probability and continuous desired outcome.  
These mixed-integer programming formulations incorporate constraints to address the desired properties required by a reliable and efficient counterfactual. Furthermore, we discuss multi-objective optimization strategies to managing the various objectives involved in the counterfactual generation.

The remainder of the paper is organized as follows. Section 2 introduces the mathematical programming framework specialized for scorecard modelling and the mixed-integer programming formulation for each outcome type. In addition, the extended formulation to impose diversity constraints is presented as well as several techniques to handle multiple objectives. Section 3 describes implementation details and includes experiments on real-world datasets to assess the quality of the generated counterfactuals and the performance of the presented approach. Finally, in Section 4, we present our conclusions.

\section{Mathematical programming framework}\label{section_mip_formulation}
In the following Section, we first introduce the general optimization framework to generating counterfactual explanations for scorecard modelling, and some properties defining a good counterfactual explanation. Secondly, we focus on mathematical programming formulations considering a scoring model taking into account the properties previously discussed.

We start considering a dataset with $n$ samples and $p$ features, where each $x_i = [x_{i1}, x_{i2}, \ldots, x_{ip}]$, and a discrete or continuous target $y_i \in \mathbb{R}$. We denote the desired outcome as $y'$ and $x'$ the counterfactual explanation. Also, we define a model $f$, with predicted outcome defined as $f(x)$.  
The objective is to search for a counterfactual $x'$ the closest to the data point $x$ such that the desired outcome $y'$ is achieved. In \cite{Wachter2018}, it is defined the problem of finding counterfactual explanations as an optimization problem. The optimization problem (\ref{counterfactual_problem}) aims to minimize the distance function, $d(x, x')$, between the input data point $x$ and the counterfactual $x'$ subject to the satisfiability of the constraint $f(x') = y'$.
\begin{subequations}\label{counterfactual_problem}
\begin{align}
\underset{x'}{\text{min}} \quad & d(x, x')\\
\text{s.t.} \quad &  f(x') = y'
\end{align}
\end{subequations}

The equality constraint in (\ref{counterfactual_problem}) might be replaced by a $l$-norm constraint given a user-defined tolerance $\epsilon$. This formulation is more adequate when the desired outcome is continuous or a class label probability.
\begin{align*}
\underset{x'}{\text{min}} \quad & d(x, x')\\
\text{s.t.} \quad &  \left|f(x') - y'\right|_l \le \epsilon
\end{align*}
Several methods solve the constrained problem (\ref{counterfactual_problem}) by reformulating it as an unconstrained problem, adding the equality constraint as a penalty in the objective function,
\begin{equation}\label{counterfactual_problem_2}
\underset{x'}{\text{min}} \; \underset{\lambda}{\text{max}} \; \lambda\left(f(x') - y'\right)^2 + d(x, x').
\end{equation}
The bilevel unconstrained problem (\ref{counterfactual_problem_2}) might be solved iteratively by increasing the penalty parameter $\lambda$ to ensure the desired outcome satisfies, within the tolerance $\epsilon$, the constraint while minimizing the distance function.

The consensus, within the machine learning explainability field, about desired properties of an effective counterfactual explanation, encompasses:
\begin{itemize}
\item Validity: a counterfactual explanation producing the desired outcome $y'$. For classification models, some methods relax this constraint by searching for a counterfactual explanation attaining an outcome as closely as possible to $y'$. This relaxation might not be appropriate to determine if a problem is infeasible, i.e., to certify that a model cannot provide a counterfactual $x'$ to change the outcome of $x$. For fairness studies, this information can be as relevant as the generated counterfactuals.
\item Proximity: a counterfactual explanation should be as close as possible to the original instance with respect to feature values. Intuitively, $x'$ becomes more doable as it approaches $x$ since fewer considerable changes are required.
\item Sparsity: a counterfactual explanation ideally should change as few features as possible to increase its understanding and effectiveness.
\item Actionability: a counterfactual must exclude value changes on non-actionable features.
\item Diversity: it is desired to generate multiple diverse counterfactual explanations to provide several alternatives to obtaining the desired outcome. This is convenient since not all individual might be equally capable of modifying the same features.

\item Data Manifold closeness or connectedness: a counterfactual explanation should be plausible, considering changes in feature values that are likely with the empirical data and correlations among variables.
\item Generation time: fast generation of one or multiple counterfactual explanations. This property is especially relevant for some applications where immediate response is required. 
\end{itemize}

\subsection{Scorecard modelling}

Scorecard modelling comprises the use of simple interpretable linear models to make predictions. Scorecard models are widely employed in sectors such as the financial industry (primarily credit risk modelling) or healthcare, requiring interpretable models with a small number of variables satisfying multiple behavioural and operational constraints. The common pipeline to develop scorecard models involves the following steps:
\begin{itemize}
\item Data processing: use of binning techniques such as optimal binning \cite{NavasPalencia2020} to discretize variables. Subsequently, apply a data transformation, generally Weight-of-Evidence (WoE) for a binary target or mean transformation for a continuous target. This step transforms the input data into binned numerical data without missing and outlier data.
\item Linear model: fit a linear model using the transformed data. Some constraints are usually incorporated, e.g., sparsity constraints are included to select only a small set of predictive features to be part of the scorecard.
\item Scorecard: the score points are calculated using the coefficients of the linear model and the data transformation assigned to each bin and variable. For example, given a feature $i$ with $\mathcal{B}_i$ bins, $j=1,\ldots, \mathcal{B}_i$, and linear model coefficient $c_i$ the corresponding score point $s_{ij}$ is given by 
\begin{equation*}
s_{ij} = c_i t_{ij},
\end{equation*}
where $t_{ij}$ is the data transformation depending on the target type. Finally, scaling methods are applied to convert score points to a common scale system.
\end{itemize}

We provide an example of a scorecard in Table \ref{table_scorecard}. The discrete nature, simplicity and relatively small size of the scorecard model make it adequate to efficiently extract optimal counterfactual explanations using combinatorial optimization techniques.

\begin{table}[H]
	\centering
	\scalebox{0.9}{
\begin{tabular}{ccc}
\hline
\textbf{Feature}    & \textbf{Bin}   & \textbf{Points} \\
\hline
ExternalRiskEstimate & [-$\infty$, 59.5) & 5.43\\
ExternalRiskEstimate & [59.5, 63.5) & 11.62\\
ExternalRiskEstimate & [63.5, 65.5) & 18.15\\
ExternalRiskEstimate & [65.5, $\infty$)  & 25.44\\
	\hline\\
	\end{tabular}}
	\caption{Example of scorecard for the feature \textit{ExternalRiskEstimate}.}
	\label{table_scorecard}
\end{table}

\subsection{Counterfactual explanation}\label{subsection_counterfactual_explanation}

In this section, we present the addressed main themes of research and desired properties of counterfactual explanations, described in the recent review \cite{Verma2020}, from a mathematical programming perspective. Then, we define three mixed-integer programming formulations for a single counterfactual explanation depending on the output type (binary, probability and continuous).

\paragraph{Proximity.} The proximity metric $d(x, x')$, measures the distance of a counterfactual $x'$ from the input data point $x$. The intuition behind this metric is that $x'$ is more realizable as it approaches $x$. Formally, it can be defined using the $l$-norm of two vectors $(x, x')$. 
\begin{equation*}
d(x, x') = \left\lVert x - x'  \right\rVert_l.
\end{equation*}

The metric $d(x, x')$ with $l=1$ or $l=2$ are generally used, but other distances such as the Huber loss can also be considered. If $l=1$, the proximity metric can be linearized obtaining a linear programming (LP) formulation. For $l=2$ and Huber loss, the proximity metric is representable using second-order cone programming (SOCP) and can be reformulated as a quadratic programming (QP).
To standardize the variability of different features, the distance metric can be weighted by the inverse median absolute deviation (MAD) or the inverse range. For $l=1$ and inverse range scaling,
\begin{equation*}
d(x, x' | w) = \sum_{i=1}^p w_i \left | x_i - x_i'  \right |,
\end{equation*}
where $w_i$ is the inverse of the value range of feature $i$, extracted from the observed data.

\paragraph{Sparsity.} Counterfactual explanations with a small number of changed features are preferred. Sparse counterfactuals are easier to understand and more achievable. The sparsity requirement can be incorporated to the objective function utilizing the term $\left\lVert x - x'\right\rVert_0$ as in \cite{Dandl_2020}, or added to the formulation as a constraint. Given the maximum number of changed features $\Theta$, and the binary variables $a_i, i=1,\ldots,p$ indicating whether a feature changed, we have
\begin{align*}\label{sparsity_constraint}
&\left\lVert x - x'\right\rVert_0  =\sum_{i=1}^p a_i, \quad \sum_{i=1}^p a_i \le \Theta.
\end{align*}

\paragraph{Actionability.} Not all model features are mutable (e.g., age, sex) or should be considered actionable (e.g., marital status). To exclude changes on these sensitive features, we add the following constraint
\begin{equation*}\label{actionability_constraint}
a_i = 0, \quad i \notin \mathcal{A},
\end{equation*}
where the set $\mathcal{A}$ contains the indices of the actionable features. The recent work in \cite{Ustun_2019} also considers features that are conditionally immutable. An example of this type of feature is the achieved education level, being a non-decreasing feature.

\paragraph{Data Manifold closeness.}

Several existing methods extract counterfactual explanations that might be considered unrealistic due to omitting the empirical data distribution such us the correlations among features. Therefore, to determine plausible counterfactuals, the empirical data distribution must be included in the objective function, so that the counterfactual are close to the training data and incorporate the observed correlations among features. 

In \cite{Kanamori_2020}, the previous considerations are tackled by adding a new term in the objective function based on the Mahalanobis distance. The Mahalanobis distance is defined for $x'\in \mathbb{R}^p$ as
\begin{equation*}
d_M(x'\,|\, \mu, \Sigma) = \sqrt{(x' - \mu)^T \Sigma^{-1}(x' - \mu)},
\end{equation*}
where $\mu$ is the estimated sample mean and $\Sigma$ is the sample covariance matrix. Given $\Sigma^{-1} \succcurlyeq 0$, $\Sigma$ can be decomposed using Cholesky decomposition as $\Sigma^{-1} = F^T F$, where $F$ is a lower triangular matrix, thus the Malahanobis distance can be rewritten as
\begin{equation*}
d_M(x'\,|\, \mu, F) = \left\lVert F(x' - \mu) \right\rVert_2.
\end{equation*}
In \cite{Kanamori_2020}, the $l_1$-norm is used instead of the $l_2$-norm to linearize the objective function,
\begin{equation*}
d_{M_1}(x' \,|\, \mu, F) = \left\lVert F(x' - \mu) \right\rVert_1.
\end{equation*}
We linearize $d_{M_1}$ using the variable splitting technique to handle the $l_1$-norm, obtaining
\begin{align*}
\underset{x'}{\text{min}} \quad & d_{M_1}(x' \,|\, \mu, F) &  \Longrightarrow & &  \underset{x'}{\text{min}} \quad & \sum_{i=1}^p \left(m_i^+ + m_i^-\right)\\
& & & & \text{s.t.} \quad &  m_i^+ - m_i^- = \sum_{j=i}^p F_{ij} (x'_j - \mu_j), & i = 1,\ldots, p\\
& & & & \quad & m_i^+, m_i^- \ge 0, & i = 1,\ldots, p
\end{align*}

There exist other robust versions of the Mahalanobis distance which could be suitable, some are discussed in \cite{Cabana2019}. Finally, other devised approaches to taking into account the relantionships among features include the computation of the weighted average distance between $x$ and the $k$ nearest observed data points \cite{Dandl_2020, Kanamori_2020}, which might be considered an empirical approximation of how likely $x$ originates from the distribution of the observed data.

\paragraph{Diversity} Some proposed algorithms are capable of generating various counterfactual explanations simultaneously for a given input data point $x$. A small set of diverse counterfactuals is useful to decide which of them are more easily achievable. Diverse is generally incorporated as a term maximizing the difference among counterfactual feature values and/or changed features \cite{Dandl_2020, Mothilal_2020}. Other algorithms \cite{Karimi2020, Ustun_2019}, based on mathematical optimization formulations, solve the same problem iteratively adding constraints to impose diversity, i.e., ensuring that new counterfactuals are substantially different from the previous ones. 

\subsubsection{Formulation for counterfactual binary outcome}
First, we consider the formulation for a binary classification problem with desired outcome $y' \in \{0, 1\}$. As previously stated, for scoring modelling we focus on  generalized linear models, which decision function is defined by
\begin{equation*}\label{decision_function}
\phi(x'| c) = \sum_{i=1}^p c_i x_i'.
\end{equation*}
Note that if the model is fitted with an intercept constant, then the decision function is given by
\begin{equation*}\label{decision_function_intercept}
\phi(x'| c) = c_0 + \sum_{i=1}^p c_i x_i'.
\end{equation*}

Let us define the parameters of the proposed mathematical programming formulation:
\begin{align*}
y' &\in \{0,1\} && \text{desired binary outcome}.\\
p &\in \mathbb{N} & & \text{number of features}.\\
x &\in \mathbb{R}^p & & \text{input data point}.\\
w &\in \mathbb{R}^p & & \text{weights to standardize feature space variability}.\\
\mathcal{B}_i & \in \mathbb{N} & & \text{number of bins per feature}.\\
woe_{ij} &\in \mathbb{R} && \text{Weight-of-Evidence per feature and bin}.\\
F & \in \mathbb{R}^{p \times (p + 1)} && \text{Cholesky decomposition of the inverse sample covariance matrix}.\\
\mu & \in \mathbb{R}^p && \text{estimated sample mean}.\\
\mathcal{A} & \in \mathbb{N}^p && \text{indices of actionable features}.\\
\Theta & \in \mathbb{N} && \text{maximum number of features to be modified}.\\
M_1, M_2 & \in \mathbb{R} && \text{minimum and maximum achievable score}.\\
\lambda_1, \lambda_2 & \in \mathbb{R}_{\ge0} && \text{weights of the objective functions}.\\
\epsilon &\in \mathbb{R}_{>0} && \text{slack parameter}.
\end{align*}

As a part of the model data preparation, a preprocessing step is devised to exclude the values of $x$ from the set of WoE values $woe_{i*}$ for each feature. Thus, $woe_{i*}$ satisfies
\begin{equation*}
\mathcal{B}_i = |woe_{i*}|, \quad x_i \notin woe_{i*}, \quad i=1,\ldots, p.
\end{equation*}
This step simplifies the formulation to detect those features that are modified, avoiding the use of $l_0$-norm terms in the objective function. The big-$M$ parameters $M_1$ and $M_2$ represent the minimum and maximum achievable score, respectively defined by
\begin{equation*}
M_1 = \sum_{i=1}^p c_i \underset{j=1\ldots,\mathcal{B}_i}{\min}\{woe_{ij}\}, \quad M_2 = \sum_{i=1}^p c_i \underset{j=1\ldots,\mathcal{B}_i}{\max}\{woe_{ij}\}
\end{equation*}

The proposed mixed-integer linear programming (MILP) formulation in detailed in (\ref{mip_binary_outcome}). The variables $t_i^+$, $t_i^-$, $m_i^+$ and $m_i^-$, and constraints (\ref{abs_d_l1}) and (\ref{abs_closeness_l1}) are part of the variable splitting linearization approach for the proximity and closeness objective function, respectively. The constraint (\ref{x_ce}) imposes that the counterfactual $x'$ must be composed by the input data point values if $z_{ij} = 0$ and actual WoE values, otherwise. The constraint (\ref{actionability_a}) forces to fix the input values for non-actionable features.

Since $y'$ is a parameter that changes the behaviour of the decision function $\phi(x'|c)$, the inequalities in (\ref{target_flip}) can be replaced by a single inequality depending on the value of $y'$. Hence,
\begin{equation*}
\begin{cases}
\phi(x'|c) > 0 \approx \phi(x'|c) \ge \epsilon, & y'=1,\\
\phi(x'|c) \le 0, & y'=0,
\end{cases}
\end{equation*}
and we use the parameter $\epsilon$, close to relative tolerance (e.g., $\epsilon = 10^{-6}$), to avoid issues with classifying the equality $\phi(x'|c) = 0$.

The parameters $\lambda_1$ and $\lambda_2$ are the weights for the proximity and closeness objective function, respectively. An objective function as a weighted sum of functions is a conventional approach to handle more than one objective function due to its simplicity. However, this method is not necessarily the most suitable for all cases. Various methods for handling multiple objective functions are discussed in Section \ref{subsection_multiobjective}.

\begin{subequations}\label{mip_binary_outcome}
\begin{align}
{\text{min}} \quad & \lambda_1\sum_{i=1}^{p} w_i \left(t_i^+ + t_i^-\right) + \lambda_2\sum_{i=1}^p \left(m_i^+ + m_i^-\right)\\
\text{s.t.} \quad &  t_i^+ - t_i^- = x_i - x'_i, & i = 1, \ldots, p\label{abs_d_l1}\\
\quad & x_i' = x_i + \sum_{j=1}^{\mathcal{B}_i} \left(woe_{ij} - x_i\right) z_{ij}, & i = 1, \ldots, p\label{x_ce}\\
\quad &  m_i^+ - m_i^- = \sum_{j=i}^p F_{ij} (x'_j - \mu_j), & i = 1,\ldots, p\label{abs_closeness_l1}\\
\quad & a_i = \sum_{j=1}^{\mathcal{B}_i} z_{ij}, & i = 1, \ldots, p\label{select_only_1_woe_1}\\
\quad & a_i \le 1, & i = 1, \ldots, p\label{select_only_1_woe_2}\\
\quad & a_i = 0, & i = 1, \ldots, p : i \notin \mathcal{A}\label{actionability_a}\\
\quad & \sum_{i=1}^p a_i \le \Theta\label{max_changes}\\
\quad & (M_1 - \epsilon) (1 - y') + \epsilon \le \phi(x' | c) \le M_2 y'\label{target_flip}\\
\quad & x_i' \in \mathbb{R}, & i = 1,\ldots, p\\
\quad & z_{ij} \in \{0, 1\}, & i = 1,\ldots, p, \; j=1,\ldots,\mathcal{B}_i\\
\quad & a_{i} \in \{0, 1\}, & i = 1,\ldots, p\\
\quad & t_i^+, t_i^- \ge 0, & i = 1,\ldots, p\\
\quad & m_i^+, m_i^- \ge 0, & i = 1,\ldots, p
\end{align}
\end{subequations}

\subsubsection{Formulation for counterfactual probability outcome}

In this case, the problem consists of adjusting the probability output $y' \in [0, 1]$, e.g., determine the changes required to reduce the probability of default from 0.6 to 0.4. The logistic regression and other generalized linear models for classification use the logistic function, 
\begin{equation}\label{logistic_function}
f(x) = \frac{1}{1 + e^{-x}},
\end{equation}
to model the probability of the event $y' = 1$. The previous MILP formulation can be extended by adding a new term in the objective function with the $l$-norm of the difference
\begin{equation}\label{logistic_function_eq}
\left\lVert y' - \frac{1}{1 + e^{-\phi(x' | c)}} \right\rVert_l
\end{equation}
or enforcing a constraint of the form
\begin{equation}\label{logistic_function_ineq}
y' \lesseqqgtr \frac{1}{1 + e^{-\phi(x' | c)}}.
\end{equation}

The incorporation of the non-convex function (\ref{logistic_function}), leads to a non-convex mixed-integer nonlinear programming (MINLP) formulation, which is significantly more challenging to solve than the previous MILP. A common approach to solving non-convex MINLP is to replace a non-convex function with a continuous piecewise linear approximation, thus obtaining a tractable MILP, at the cost of enlarging the resulting formulation. 
We approximate the logistic function using a continuous piecewise linear function with $R$ segments with coefficients $(\beta_r, \alpha_r)$ for $r=1, \ldots, R$ such that
\begin{equation*}
f(x) \approx \tilde{f}(x) = \alpha_r x + \beta_r, \quad x \in [b^{r-1}, b^r], \quad r = 1, \ldots, R,
\end{equation*}
where the pair of break points $[b^{r-1}, b^r], r=1,\ldots, R$ form the interval of each segment. The piecewise linear approximation is obtained using $R-1$ breakpoints in the interval $[M_1, M_2]$ satisfying 
\begin{equation*}
\max_{x\in [M_1, M_2]} \left | f(x) - \tilde{f}(x)\right | < \epsilon_{approx}.
\end{equation*}
Note that there is a trade-off between accuracy and the number of segments $R$ since as $R$ increases the approximation improves, but the problem size increases, possibly worsening solution times. There are several strategies to select the break points $b^r$. Common strategies are selecting break points uniformly in a given interval, or a Greedy approach where the set of break points is incrementally constructed by adding a new break point in the segment with the largest approximation error.

The resulting formulation after applying the discussed linearization is given by
\begin{subequations}
\begin{align}
{\text{min}} \quad & \begin{aligned}[t]
& \lambda_1\sum_{i=1}^{p} w_i \left(t_i^+ + t_i^-\right) + \lambda_2 \sum_{i=1}^p \left(m_i^+ + m_i^-\right)\\ &+ \lambda_3 \left(q^+ + q^-\right)\label{abs_logistic_dl1}\\
\end{aligned}\\
\text{s.t.} \quad & \text{(\ref{abs_d_l1} - \ref{max_changes})}\\
\quad & \sum_{r=1}^R h_r = \phi(x' | c)\label{piecewise_0}\\
\quad & b^{r-1} s_r \le h_r \le b^r s_r, & r=1, \ldots, R\label{piecewise_1}\\
\quad & \sum_{r=1}^R \left(\alpha_r h_r + \beta_r s_r\right) = f\label{piecewise_2}\\
\quad & \sum_{r=1}^R s_r = 1\label{piecewise_3}\\
\quad & q^+ - q^- = f - y'\label{abs_logistic_constr}\\
\quad & x_i' \in \mathbb{R}, & i = 1,\ldots, p\\
\quad & z_{ij} \in \{0, 1\}, & i = 1,\ldots, p; \; j=1,\ldots,\mathcal{B}_i\\
\quad & a_{i} \in \{0, 1\}, & i = 1,\ldots, p\\
\quad & t_i^+, t_i^- \ge 0, & i = 1,\ldots, p\\
\quad & h_r \in \mathbb{R}, & r = 1,\ldots, R\\
\quad & s_r \in \{0, 1\}, & r = 1,\ldots, R\\
\quad & q^+, q^-, f \ge 0
\end{align}
\end{subequations}

Now, we describe the main differences with respect to the formulation for a binary outcome. First, the third term in (\ref{abs_logistic_dl1}), $\lambda_3\left(q^+ + q^-\right)$, and the constraint (\ref{abs_logistic_constr}) represent the linearization of (\ref{logistic_function_eq}) when $l=1$. Second, the constraints (\ref{piecewise_0} - \ref{piecewise_3}) is the piecewise linear approximation of the logistic function (\ref{logistic_function}). Moreover, as previously discussed, the third term in the objective function can be replaced by the constraint (\ref{logistic_function_ineq}). This can be accomplished by adding the constraint $f \le y'$ or $f \ge y'$ and removing the third term from the objective. Also, using the previous modification, the auxiliary variables $q^+$ and $q^-$ and the constraint (\ref{abs_logistic_constr}) are no longer needed.

\subsubsection{Formulation for counterfactual continuous outcome}
In the case of a continuous target, the problem consists of approximating the desired output $y'$.  Using the linear model $\phi(x' | c)$, the following term is included in the objective function,
\begin{equation}\label{continuous_obj}
\lVert y' - \phi(x' | c)\rVert_l, \quad l \in \{1, 2\}.
\end{equation}
Besides, the formulation might be complemented with an additional constraint added to enforce a minimum or maximum desired output,
\begin{equation*}
y' \ge \phi(x' | c), \quad y' \le \phi(x' | c).
\end{equation*}
Note that, as described in the probability outcome case, the $l$-norm term can also be replaced completely by one of the above constraints. Given the discrete search space of $x'$, the equality constraint $y' = \phi(x' | c)$ is not contemplated since it could often lead to an infeasible formulation.

We propose a MILP formulation of the form:
\begin{subequations}
\begin{align}
{\text{min}} \quad & \begin{aligned}[t]
& \lambda_1\sum_{i=1}^{p} w_i \left(t_i^+ + t_i^-\right) + \lambda_2 \sum_{i=1}^p \left(m_i^+ + m_i^-\right)\\ &+ \lambda_3 \left(q^+ + q^-\right)\\
\end{aligned}\\
\text{s.t.} \quad &  t_i^+ - t_i^- = x_i - x'_i, & i = 1, \ldots, p\\
\quad & x_i' = x_i + \sum_{j=1}^{\mathcal{B}_i} \left(mean_{ij} - x_i\right) z_{ij}, & i = 1, \ldots, p\\
\quad &  m_i^+ - m_i^- = \sum_{j=i}^p F_{ij} (x'_j - \mu_j), & i = 1,\ldots, p\\
\quad & a_i = \sum_{j=1}^{\mathcal{B}_i} z_{ij}, & i = 1, \ldots, p\\
\quad & a_i \le 1, & i = 1, \ldots, p\\
\quad & a_i = 0, & i = 1, \ldots, p : i \notin \mathcal{A}\\
\quad & \sum_{i=1}^p a_i \le \Theta\\
\quad & q^+ - q^- = \phi(x' | c) - y'\label{diff_continuous_obj}\\
\quad & x_i' \in \mathbb{R}, & i = 1,\ldots, p\\
\quad & z_{ij} \in \{0, 1\}, & i = 1,\ldots, p, \; j=1,\ldots,\mathcal{B}_i\\
\quad & a_{i} \in \{0, 1\}, & i = 1,\ldots, p\\
\quad & t_i^+, t_i^- \ge 0, & i = 1,\ldots, p\\
\quad & m_i^+, m_i^- \ge 0, & i = 1,\ldots, p\\
\quad & q^+, q^-, f \ge 0
\end{align}
\end{subequations}

Here, the WoE transformation $woe_{ij}$ are replaced by the mean transformation $mean_{ij}$, and constraint (\ref{diff_continuous_obj}) expresses the linearization of the term (\ref{continuous_obj}) when selecting the $l_1$-norm.

\subsection{Multiple counterfactual explanations}\label{multiple_cfs}

In this section, we present a mathematical programming formulation to generating multiple counterfactual explanations for a single data point simultaneously. In some recent works \cite{Ustun_2019, Karimi2020}, multiple counterfactual explanations for a single data point have been computed iteratively. At each iteration, these methods solve an optimization problem adding constraints to enforce diversity. Contrarily, we aim to generate $K$ counterfactual simultaneously, while maximizing or enforcing diversity regarding the changed features and the values of the features, solving a single optimization problem.

In the following, we choose a binary desired outcome $y'$, but this approach is extensible to other types of counterfactual outcomes. We start treating diversity as hard constraints. Then, we include them in the objective function to avoid infeasibility issues when there are fewer features or sparsity constraints are also imposed.

\subsubsection{Diversity constraints}

First, let us focus on the diversity of features to be changed. We consider the indicator variable $a_i$, and two counterfactuals $k$ and $l$. The difference between $a_{ki}$ and $a_{li}$ denoted as $u_{kli}$ can be calculated using the absolute distance, equivalent to the XOR operator:
\begin{equation}\label{diversity_xor}
u_{kli} = a_{ki} \oplus a_{li} = |a_{ki} - a_{li}|.
\end{equation}
The XOR in (\ref{diversity_xor}) can be linearized as follows:
\begin{align*}
u_{kli} &\le a_{ki} + a_{li}\\
u_{kli} &\ge a_{ki} - a_{li}\\
u_{kli} &\ge - a_{ki} + a_{li}\\
u_{kli} &\le 2 - a_{ki} - a_{li}
\end{align*}
for $k=1,\ldots,K;\; l=k+1,\ldots,K; \; i=1,\ldots,p$. To enforce a different combinations of features for all counterfactual explanations we add the following constraint
\begin{equation}\label{diversity_features}
\sum_{i=1}^p u_{kli} \ge 1, \quad k=1,\ldots,K;\; l=k+1,\ldots,K.
\end{equation}

The diversity of feature values can also be addressed by adding additional constraints. We consider the indicator variable $z_{ij}$, indicating the new value index for feature $j$ if changed, and two counterfactuals $k$ and $l$. The difference between the feature values between two counterfactuals can be expressed as previously
\begin{equation*}
d_{klij} = z_{kij} \oplus z_{lij} = |z_{kij} - z_{lij}|,
\end{equation*}
which can be linearized as (\ref{diversity_xor})
\begin{align*}
d_{klij} &\le z_{kij} + z_{lij}\\
d_{klij} &\ge z_{kij} - z_{lij}\\
d_{klij} &\ge - z_{kij} + z_{lij}\\
d_{klij} &\le 2 - z_{kij} - z_{lij}
\end{align*}
for $k=1,\ldots,K;\; l=k+1,\ldots,K; \; i=1,\ldots,p; \; \forall j \in \mathcal{B}_i$. To ensure that counterfactual explanations are generated without repeated values for a changed feature we use the constraint
\begin{equation}\label{diversity_feature_values}
\sum_{j=1}^{\mathcal{B}_i}d_{klij} \ge a_{ki} + a_{li} - 1,  \quad k=1,\ldots,K;\; l=k+1,\ldots,K;\; i=1,\ldots,p.
\end{equation}
Note that this can be formulated using implication constraints. Only enforce if the counterfactual explanations changed the same feature: $a_{ki} = a_{li} = 1$:
\begin{equation*}
a_{ki} \land a_{li} \Rightarrow \sum_{j=1}^{\mathcal{B}_i}d_{klij} \ge 1 \Longleftrightarrow \sum_{j=1}^{\mathcal{B}_i}d_{klij} \ge a_{ki} + a_{li} - 1.
\end{equation*}

The diversity constraints are incorporated to the formulation devised in (\ref{mip_binary_outcome}), extended to generating $K$ counterfactuals, obtaining
\begin{subequations}\label{mip_diversity_binary_outcome}
\begin{align}
{\text{min}} \quad & \sum_{k=1}^K \left(\lambda_1\sum_{i=1}^{p} w_i \left(t_{ki}^+ + t_{ki}^-\right) +  \lambda_2\sum_{i=1}^p \left(m_i^+ + m_i^-\right)\right)\label{obj_multiple}\\
\text{s.t.} \quad &  t_{ki}^+ - t_{ki}^- = x_i - x'_{ki}, & k=1,\ldots, K; \; i = 1, \ldots, p\\
\quad & x_{ki}' = x_i + \sum_{j=1}^{\mathcal{B}_i} \left(woe_{ij} - x_{ki}\right) z_{kij}, & k=1,\ldots, K; \; i = 1, \ldots, p\\
\quad & (m - \epsilon) (1 - y') + \epsilon \le \sum_{i=1}^p c_i x_{ki}', & k=1,\ldots, K\\
\quad & \sum_{i=1}^p c_i x_{ki}' \le M y', & k=1,\ldots, K\\
\quad & a_{ki} = \sum_{j=1}^{\mathcal{B}_i} z_{kij}, &  k=1,\ldots, K; \; i = 1, \ldots, p\\
\quad & a_{ki} \le 1, &  k=1,\ldots, K; \; i = 1, \ldots, p\\
\quad & \sum_{i=1}^p a_{ki} \le \Theta, & k=1,\ldots, K\\
\quad & (\ref{diversity_features}-\ref{diversity_feature_values})\\
\quad & x_i' \in \mathbb{R}, & k=1,\ldots, K; \; i = 1, \ldots, p\\
\quad & z_{kij} \in \{0, 1\}, & k=1,\ldots, K; \; i = 1,\ldots, p; \; \forall j \in \mathcal{B}_i\\
\quad & a_{ki} \in \{0, 1\}, & k=1,\ldots, K; \;i = 1,\ldots, p\\
\quad & t_{ki}^+, t_{ki}^- \ge 0, & k=1,\ldots, K; \;i = 1,\ldots, p
\end{align}
\end{subequations}

\subsubsection{Diversity objective}
To compute the distance between two counterfactuals, we use the fact that the scorecard is discrete and a counterfactual can be fully characterize by the binary decision variables $z_{ij}$. Given two binary arrays, $x, y \in \{0, 1\}^p$, their distance can be calculated using the Hamming distance given by
\begin{equation*}
d_H(x, y) = \sum_{i=1}^p \mathbf{1}_{\{x_i \neq y_i\}} = \sum_{i=1}^p x_i \oplus y_i.
\end{equation*}

Thus, we define the diversity of changed features as the sum of elements in the lower triangular pairwise Hamming distance matrix among counterfactuals $a_i$, $i = 1,\ldots, n$:
\begin{equation*}
D_{F} = \sum_{k=1}^K \sum_{l=k+1}^K \sum_{i=1}^p u_{kli}
\end{equation*}
Analogously, we define the diversity of feature values among counterfactuals using $z_{ij}$, $i = 1,\ldots, p; \; \forall j \in \mathcal{B}_i$ as follows
\begin{equation*}
D_{FV} = \sum_{k=1}^K \sum_{l=k+1}^K \sum_{i=1}^p \sum_{j=1}^{\mathcal{B}_i} d_{klij}.
\end{equation*}

Finally, the objective (\ref{obj_multiple}) is updated taking into account that diversity objectives are maximized and assigning their corresponding weights
\begin{equation*}
{\text{min}}\;\sum_{k=1}^K \left(\lambda_1\sum_{i=1}^{p} w_i \left(t_{ki}^+ + t_{ki}^-\right) +  \lambda_2\sum_{i=1}^p \left(m_i^+ + m_i^-\right) -\lambda_3 \sum_{l=k+1}^K \sum_{i=1}^p u_{kli} -\lambda_4 \sum_{l=k+1}^K \sum_{i=1}^p \sum_{j=1}^{\mathcal{B}_i} d_{klij} \right).
\end{equation*}

\subsection{Multi-objective counterfactual explanation}\label{subsection_multiobjective}

In all formulations from Sections \ref{subsection_counterfactual_explanation} and \ref{multiple_cfs}, the various objectives functions are combined creating a single objective as a linear combination with given weights $\lambda$. This is a widespread approach for multi-objective optimization known as weighted or blended objective. However, assigning the appropriate weights to balance various competing objectives is challenging. Also, objectives with different scales require a thoughtful selection of weights to choose the correct relative importance of each objective.

Another multi-objective strategy implemented in various commercial optimization solvers is the hierarchical or lexicographic approach. This approach requires determining the priority order of the objectives and optimizes sequentially in decreasing order. In practice, higher priority objectives are allowed to be degradable by a small deviation. More precisely, objective degradations are handled by adding extra constraints imposing a maximum absolute or relative deviation with respect to the optimal solution using a particular objective. The main disadvantage compared to the blended approach, is the computational cost of solving the optimization problem for each objective.

Recently, the counterfactual explanation algorithm described in \cite{Dandl_2020}, proposed the use  of the evolutionary strategy \textit{Nondominated Sorting Genetic Algorithm II} (NSGA II) \cite{Deb2002} to determine the Pareto frontier, i.e., the boundary defined by the feasible non-dominated solutions representing different trade-offs among objectives.

Our focus is on extracting optimal or good feasible counterfactuals quickly, hence we primarily investigate the performance of the blended and hierarchical approach in Section \ref{section_experiments}. Finally, we discard genetic algorithms searching the Pareto frontier due to the computational cost.

\section{Experiments}\label{section_experiments}

\subsection{Implementation}
The mathematical programming formulations are implemented using Google OR-Tools \cite{ortools} with the open-source MILP solver CBC \cite{cbc} and Google's CP-SAT solver. Furthermore, for solving multi-objective optimization problems using Google OR-Tools, we implemented a custom hierarchical approach, thus supporting two strategies to generating counterfactual explanations. The scorecard models are developed using the OptBinning library \cite{NavasPalencia2020}, freely available\footnote{\href{https://github.com/guillermo-navas-palencia/optbinning}{https://github.com/guillermo-navas-palencia/optbinning}}. The logistic regression in Scikit-learn \cite{scikit-learn} with $l_2$ regularization is used as the estimator of the scorecard model. Finally, the implementation of counterfactual explanations in OptBinning will be available in the release 0.11.0.

\subsection{Experiment: binary target}

To evaluate the performance and quality of the counterfactual explanations generated using the presented approach, we consider two datasets:
\begin{itemize}
\item \textbf{FICO}: This is an anonymized dataset from the FICO Explainable Machine Learning Challenge \cite{FICO2018}. This dataset contains real Home Equity Line of Credit (HELOC) applications. The task is to predict whether individuals will repay their HELOC account within 2 years.

\item \textbf{Adult-Income}: This dataset contains information based on 1994 Census database \cite{Adult1996}. We perform the data processing described in \cite{Mothilal_2020}, selecting the same 8 features, namely, \textit{hours per week, education level, occupation, work class, race, age, marital status}, and \textit{sex}. The task is to classify whether an individual’s income exceeds $\$50$K/year.
\end{itemize}

For both datasets, a scorecard model is developed using the \texttt{Scorecard} class in OptBinning\footnote{Tutorial step by step: \href{http://gnpalencia.org/optbinning/tutorials/tutorial_scorecard_binary_target.html}{http://gnpalencia.org/optbinning/tutorials/tutorial\_scorecard\_binary\_target.html}}. The experiments were run on an Intel(R) Core(TM) i5-3317 CPU at 1.70GHz running Linux. Both solvers, CBC and CP-SAT, use 1 thread.

\subsubsection{Single counterfactual}
In this first set of experiments, we use the FICO dataset considering the 12 features selected by the scorecard model. We first compare the counterfactual explanations generated using the weighted and hierarchical approach to handle several objectives. Table \ref{table_cf_1_weighted} reports generated counterfactuals using the weighted approach with weights $\lambda_1 = \lambda_2 = 1$. For each counterfactual a different sparsity limit is set, $\Theta \in \{1, 2, 3, 4\}$.
\begin{table}[H]
	\centering
	\scalebox{0.9}{
\begin{tabular}{ccc}
\hline
\textbf{Feature}    & \textbf{Current value}   & \textbf{Required value} \\
\hline
PercentTradesNeverDelq & 83 & [97.50, $\infty$)\\
\hline
PercentTradesNeverDelq & 83 & [91.50, 95.50)\\
MSinceMostRecentInqexcl7days & 0  & [1.50, 10.50)\\
\hline
PercentTradesNeverDelq & 83 & [91.50, 95.50)\\
MSinceMostRecentInqexcl7days & 0  & [0.50, 1.50)\\
NumBank2NatlTradesWHighUtilization & 2 & [0.50, 1.50)\\
\hline
PercentTradesNeverDelq & 83 & [91.50, 95.50)\\
MSinceMostRecentInqexcl7days & 0  & [0.50, 1.50)\\
NetFractionRevolvingBurden & 28 & [31.50, 37.50)\\
NumBank2NatlTradesWHighUtilization & 2 & [0.50, 1.50)\\
	\hline\\
	\end{tabular}}
	\caption{A single counterfactual explanation using $\Theta \in \{1, 2, 3, 4\}$ and the weighted approach.}
	\label{table_cf_1_weighted}
\end{table}

Table \ref{table_cf_1_hierarchical} shows the generated counterfactuals using the hierarchical with maximum relative degradation of the first objective set to 0.1, and sparsity $\Theta = 2$. The first counterfactual prioritizes the proximity objective, whereas the second prioritizes the closeness requirement. Note that when proximity is prioritized, the feature AverageMInFile replaces MSinceMostRecentInqexcl7days since the relative difference in MSinceMostRecentInqexcl7days is greater than the corresponding for AverageMInFile.
\begin{table}[H]
	\centering
	\scalebox{0.9}{
\begin{tabular}{ccc}
\hline
\textbf{Feature}    & \textbf{Current value}   & \textbf{Required value} \\
\hline
AverageMInFile & 65 & [97.50, 116.50)\\
PercentTradesNeverDelq & 83 & [91.50, 95.50)\\
\hline
PercentTradesNeverDelq & 83 & [88.50, 91.50)\\
MSinceMostRecentInqexcl7days & 0  & [1.50, 10.50)\\
	\hline\\
	\end{tabular}}
	\caption{A single counterfactual explanation using $\Theta=2$ and the hierarchical approach.}
	\label{table_cf_1_hierarchical}
\end{table}

As mentioned in Section \ref{subsection_multiobjective}, solving a multi-objective optimization problem using the hierarchical approach might increase the computation time. Table \ref{table_cf_1_objectives_approach} compares both approaches in terms of the objective functions and the CPU times using the CBC solver, confirming that with two objectives the CPU approximately doubles. Besides, given the probability of default (PD) for this particular data point is 0.742, column PD shows the PD assigned by the scorecard model to the generated counterfactual.

\begin{table}[H]
	\centering
	\scalebox{0.9}{
\begin{tabular}{cccccc}
\hline
\boldmath$\Theta$  & \textbf{Approach}  & \textbf{Proximity}   & \textbf{Closeness}  & \textbf{PD} & \textbf{Time (s)}\\
\hline
1 & W & 1.134 & 9.356 & 0.484 & 0.3\\
2 & W & 1.672 & 7.544 & 0.436 & 0.3\\
3 & W & 1.767 & 5.524 & 0.478 & 0.4\\
4 & W & 1.816 & 5.406 & 0.489 & 0.5 \\
\hline
2 & H(1, 0) & 0.993 & 8.742 & 0.489 & 0.5\\
2 & H(0, 1) & 1.504 & 8.206 & 0.476 & 0.9\\
	\hline\\
	\end{tabular}}
	\caption{Performance and counterfactual PD comparison weighted (W) vs hierarchical (H) approach. H(1, 0) and H(0, 1) prioritize proximity and closeness, respectively.}
	\label{table_cf_1_objectives_approach}
\end{table}

\subsubsection{Multiple counterfactuals}
The experiments in this section serve to evaluate the quality of the generated counterfactuals when incorporating diversity constraints. First, we use the same data point in Table \ref{table_cf_1_weighted} to generate three counterfactuals using the diversity constraint regarding the values of the features in (\ref{diversity_feature_values}). Note that the second counterfactual in Table \ref{table_mcf_weighted} is the counterfactual in Table \ref{table_cf_1_weighted} when $\Theta = 2$. We can observe that the combination of features is not unique, but no feature value is repeated on the changed features.
\begin{table}[H]
	\centering
	\scalebox{0.9}{
\begin{tabular}{ccc}
\hline
\textbf{Feature}    & \textbf{Current value}   & \textbf{Required value} \\
\hline
PercentTradesNeverDelq & 83 & [97.50, $\infty$)\\
NumBank2NatlTradesWHighUtilization & 2 & [0.50, 1.50)\\
\hline
PercentTradesNeverDelq & 83 & [91.50, 95.50)\\
MSinceMostRecentInqexcl7days & 0 & [1.50, 10.50)\\
\hline
PercentTradesNeverDelq & 83 & [95.50, 97.50)\\
MSinceMostRecentInqexcl7days & 0 & [0.50, 1.50)\\
	\hline\\
	\end{tabular}}
	\caption{Three counterfactual explanations using $\Theta = 2$ and the weighted approach. Feature values as hard constraint. Feature changes not imposed.}
	\label{table_mcf_weighted}
\end{table}

The hierarchical approach can also be used to generate multiple counterfactuals. Table \ref{table_mcf_objectives_approach} compares both approaches reporting the average proximity and closeness, and the diversity metrics $D_F$ and $D_{FV}$. In addition, the minimum and maximum PD among the generated counterfactuals is reported. These result may vary considerably by modifying the weights $\lambda$ and the maximum relative degradation.

\begin{table}[H]
	\centering
	\scalebox{0.9}{
\begin{tabular}{ccccccccc}
\hline
\boldmath$K$ & \boldmath$\Theta$  & \textbf{Approach}  & \textbf{Proximity}   & \textbf{Closeness} & \boldmath$D_F$ & \boldmath$D_{FV}$ & \textbf{PD$_{\min}$} & \textbf{PD$_{\max}$}\\
\hline
3 & 2 & W & 1.638 & 7.901 & 4 & 12 & 0.422 & 0.457\\
4 & 3 & W & 1.807 & 7.494 & 16 & 36 & 0.455 & 0.487\\
3 & 2 & H(1, 0) & 1.096 & 8.895 & 4 & 12 & 0.477 & 0.495\\
3 & 2 & H(0, 1) & 1.203 & 8.561 & 4 & 12 & 0.477 & 0.489\\
	\hline\\
	\end{tabular}}
	\caption{Metrics of multiple counterfactual explanations for $K \in \{3, 4\}$, $\Theta \in \{2, 3\}$, using the weighted (W) and hierarchical (H) approach. H(1, 0) and H(0, 1) prioritize proximity and closeness, respectively.}
	\label{table_mcf_objectives_approach}
\end{table}

For the remaining experiments we use the Adult-Income dataset. We take the data point chosen in \cite{Mothilal_2020} to conduct experiments. Table \ref{table_mcf_mip_weighted} shows four counterfactuals generated with sparsity $\Theta = 4$, imposing both diversity constraints and using the weighted approach.
\begin{table}[H]
	\centering
	\scalebox{0.9}{
\begin{tabular}{ccc}
\hline
\textbf{Feature}    & \textbf{Current value}   & \textbf{Required value} \\
\hline
age & 22 & [43.50, 49.50)\\
education & HS-grad & Bachelors\\
\hline
age & 22 & [35.50, 37.50)\\
education & HS-grad & [Masters, Prof-school, Doctorate]\\
hours-per-week & 45 &  [39.50, 41.50)\\
\hline
age & 22 & [49.50, 54.50)\\
education & HS-grad & Some-college\\
marital-status & Single & [Married-AF-spouse, Married-civ-spouse]\\
\hline
age & 22 & [40.50, 43.50)\\
education & HS-grad & [Assoc-acdm, Assoc-voc]\\
hours-per-week & 45 & [55.50, $\infty$)\\
occupation & Service & Exec-managerial\\
\hline
	\end{tabular}}
	\caption{Four counterfactual explanations using $\Theta = 4$ and the weighted approach. Feature changes and feature values as hard constraint.}
	\label{table_mcf_mip_weighted}
\end{table}

We note that unlike the DiCE method in \cite{Mothilal_2020}, our method includes both proximity and closeness objectives. The main observations from Table \ref{table_mcf_mip_weighted} are that neither feature \textit{sex} nor \textit{race} is included, a higher salary requires an advanced degree which in turn requires years of study, showing the underlying positive correlation among these features. Excluding the closeness objective, we got the four counterfactuals in Table \ref{table_mcf_cp_weighted}.
\begin{table}[H]
	\centering
	\scalebox{0.9}{
\begin{tabular}{ccc}
\hline
\textbf{Feature}    & \textbf{Current value}   & \textbf{Required value} \\
\hline
age & 22 & [31.50, 33.50)\\
education & HS-grad & [Masters, Prof-school, Doctorate]\\
\hline
age & 22 & [37.50, 40.50)\\
education & HS-grad & Bachelors\\
hours-per-week & 45 &  [55.50, inf)\\
\hline
age & 22 & [49.50, 54.50)\\
education & HS-grad & Some-college\\
marital-status & Single & [Married-AF-spouse, Married-civ-spouse]\\
\hline
age & 22 & [43.50, 49.50)\\
education & HS-grad & [Assoc-acdm, Assoc-voc]\\
occupation & Service & Exec-managerial\\
\hline
	\end{tabular}}
	\caption{Four counterfactual explanations using $\Theta = 4$ and the weighted approach. Closeness objective not included. Feature changes and feature values as hard constraint.}
	\label{table_mcf_cp_weighted}
\end{table}

Finally, we report CPU times using the CP-SAT solver in Table \ref{table_mcf_cp_performance}. In our experiments, CP-SAT is significantly faster than CBC as $K$ increases, being more noticeable when both diversity constraints are considered.
\begin{table}[H]
	\centering
	\scalebox{0.9}{
\begin{tabular}{cccccc}
\hline
\boldmath$K$ & \boldmath$\Theta$ & \textbf{Diversity} & \textbf{P[$> 50K$]$_{\min}$} & \textbf{P[$> 50K$]$_{\max}$} & \textbf{Time (s)}\\
\hline
3 & 2 & F + FV & - & - & infeasible (0.1)\\
3 & 3 & F + FV & 0.506 & 0.510 & 0.6\\
3 & 4 & F + FV & 0.506 & 0.510 & 3.9\\
4 & 4 & F + FV & 0.502 & 0.521 & 2.5\\
5 & 3 & F + FV & - & - & infeasible (0.8)\\
5 & 4 & F + FV & - & - & infeasible (0.4)\\
5 & 3 & F & 0.503 & 0.526 & 5.1\\
5 & 4 & F & 0.503 & 0.526 & 13.7\\
5 & 5 & F & 0.502 & 0.522 & 9.7\\
5 & 5 & F + NA & 0.502 & 0.526 & 21.3\\
	\hline\\
	\end{tabular}}
	\caption{Metrics of multiple counterfactual explanations for $K \in \{3, 4, 5\}$, $\Theta \in \{2, 3, 4, 5\}$, using the weighted approach and CP-SAT solver. F: feature changes. FV: feature values.  NA: features \textit{marital-status, sex} and \textit{race} are non-actionable.}
	\label{table_mcf_cp_performance}
\end{table}

Table \ref{table_mcf_cp_performance} shows that multiple counterfactuals can be generated simultaneously for a single data point in a few seconds. Besides, this approach is capable of providing a certificate of infeasibility quickly, confirming that no counterfactual can be generated by the scorecard model under these constraints. If the number of actionable features is small and the number of bins is limited, one can easily reach infeasible solutions if the diversity constraint regarding feature values is enforced. Therefore, a granular scorecard is beneficial to find several diverse counterfactuals.

For very large problems, obtaining optimal solutions might take minutes, but we found that good feasible solutions are generally achieved within seconds. Thus, a practical implementation would require a time limit parameter.

\section{Conclusions}\label{section_conclusions}

We proposed several mathematical programming formulations to generating diverse counterfactual explanations. The presented method demonstrates that optimal counterfactuals with diversity constraints are quickly generated for a scorecard model. Also, these formulations can easily incorporate additional constraints such as causal relationships and allowed ranges for feature values.

Solving an optimization problem for each data point in the dataset can be time-consuming. A reasonable extension of this work is the development of fast heuristic methods to generate feasible counterfactuals, thus enabling a continuous assessment of the scorecard model biases. Furthermore, this would allow performing a careful evaluation of a scorecard model before putting it into production.

\bibliographystyle{plain}

\end{document}